\documentclass[runningheads]{llncs}
\usepackage[T1]{fontenc}
\usepackage{graphicx}
\usepackage{capt-of}  
\usepackage{cuted}
\usepackage[backend=biber, sorting=none, giveninits=true]{biblatex}
\usepackage{graphicx}%
\usepackage{multirow}%
\usepackage{amsmath,amssymb,amsfonts}%
\usepackage{mathrsfs}%
\usepackage[figuresright]{rotating}%
\usepackage{xcolor}%
\usepackage{textcomp}%
\usepackage{manyfoot}%
\usepackage{booktabs}%
\usepackage{algorithm}%
\usepackage{algorithmicx}%
\usepackage{algpseudocode}%
\usepackage{program}%
\usepackage{listings}%
\usepackage{wrapfig}
\usepackage{vruler}
\usepackage{setspace}
\usepackage{comment}
\usepackage{siunitx}
\usepackage{booktabs}

\begin{document}
\title{On the Computation of Meaning, Language Models and Incomprehensible Horrors}
\titlerunning{On the Computation of Meaning}
%
%
\author{Michael Timothy Bennett\inst{1} \\\orcidID{0000-0001-6895-8782} 
}
\authorrunning{Michael Timothy Bennett}
%
\institute{The Australian National University \\
\email{michael.bennett@anu.edu.au} \\
\url{http://www.michaeltimothybennett.com/}
}
\maketitle              
%


\begin{abstract}We integrate foundational theories of meaning with a mathematical formalism of artificial general intelligence (AGI) to offer a comprehensive mechanistic explanation of meaning, communication, and symbol emergence. This synthesis holds significance for both AGI and broader debates concerning the nature of language, as it unifies pragmatics, logical truth conditional semantics, Peircean semiotics, and a computable model of enactive cognition, addressing phenomena that have traditionally evaded mechanistic explanation. By examining the conditions under which a machine can generate meaningful utterances or comprehend human meaning, we establish that the current generation of language models do not possess the same understanding of meaning as humans nor intend any meaning that we might attribute to their responses. To address this, we propose simulating human feelings and optimising models to construct weak representations. Our findings shed light on the relationship between meaning and intelligence, and how we can build machines that comprehend and intend meaning\footnote{Appendices are to be found on GitHub \cite{bennett2023appendices}.}.
\end{abstract}

\section{Introduction}
\label{intro}

Linguists and philosophers have offered various accounts of the behaviour of language and the human mind. Computer scientists have posited mechanisms to replicate these variously described behaviours piecemeal. The former is a top-down approach, while the latter is bottom up.
Unfortunately, it is difficult to connect the two. Large language models (LLMs) such as ChatGPT are a bottom up attempt to capture the behaviour of written language, and are remarkably good at giving human-like responses to questions. Yet it is unclear the extent to which an LLM actually means what it says or understands what we mean.
AGI should not just parrot what we expect but respond to what we mean, and mean what it says. Yet how we would we know the difference? Computers represent syntax, and from correlations in syntax an LLM is supposed to glean meaning. However, meaning is not well defined. We need to connect top-down descriptions of meaning to bottom-up computation. How might we compute meaning?

\subsection{Grice's foundational theory of meaning}
Grice's foundational theory of meaning \cite{grice1957} holds that meaning is what the speaker \textit{intends} to convey to the listener. Grice gave an illustrative example,  
\begin{quote}
[the speaker] $\alpha$ means $m$ by uttering $u$ iff $\alpha$ intends in uttering $u$ that \\
1. his audience come to believe $m$, \\
2. his audience recognize this intention [called m-intention], and \\
3. (1) occur on the basis of (2). \cite{speaks2021}
\end{quote}
This is \textit{foundational} because it specifies the facts in virtue of which expressions have particular semantic properties (instead of those properties), and is illustrative of our goal (to connect bottom up computation to top down description).

\subsection{A foundational theory of foundational theories}
Were we to accept that meaning is in virtue of m-intent\footnote{We note that Grice later expanded upon the notion of m-intent \cite{grice1969,grice2007}, and that there are other widely accepted descriptions of meaning (Russell, Frege, Searle, Davidson, Wittgenstein, Lewis, Kripke etc), some of which we touch upon as part of our formalism. However, paper length limits what we discuss. }
, then from what does that arise? M-intent should not be conflated with intent in general because it pertains to what one means by an expression, whereas intent more generally is any goal in service of which decisions are made. The former stems from the latter \cite{bennett2022a}, and so there exists a theory arguing that meaning exists in virtue of one's intent in the sense of goals. Grice's theories are better established and widely accepted with respect to meaning, but these theories are not mutually exclusive and the depiction of meaning as in virtue of intent in general is a bridge we can use to connect Grice's top down description to bottom-up computational processes. This is because it explains intent in virtue of inductive inference, to argue that meaningful communication with an AI, or any organism, requires similar feelings and experiences, in order to construct similar goals and ``solutions to tasks'' \cite{bennett2022a} (an argument formed in relation to the Fermi Paradox \cite{bennett2022b}). This explanation was too vague to be of significance for engineering. For example it assumed a measure, ``weakness'', which was not well defined. However, weakness \textit{is} well defined in a more recent formalism of artificial general intelligence (AGI) \cite{bennett2022c,bennett2023appendices} and enactive cognition, so we will instead reformulate the theory using that formalism, extending it to account for meaningful communication. 
We begin with cognition formalised using tasks. We then formalise an organism using tasks to provide a novel account of preferences, symbol systems and meaningful communication. We then describe how an organism might mean what we think it means by what it says, or infer what we mean by what we say. 

\section{Meaning, from the top down}
Intent only exists in virtue of a task one is undertaking \cite{bennett2022a}. A task is what we get if we add context to intent, expressing what is relevant about both the agent\newpage \noindent and the environment. A task can be used to formalise enactive cognition  \cite{ward2017}, discarding notions of agent and environment in favour of a set of decision problems \cite{bennett2022a,bennett2023appendices}.
A task is something which is completed, like a goal, so intent is formalised like a goal \cite{bennettmaruyama2022a}. A goal is a set of criteria, and if those criteria are satisfied, then it is satisfied and the task complete. To formalise meaning we must avoid grounding problems \cite{harnad1990}. As such these criteria are grounded by representing the environment, of which cognition is part, as a set of declarative programs \cite{howard1980} of which the universe is the interpreter \cite{piccinini2021}:  
\begin{definition}[environment]\label{d1}
\begin{itemize}{
    \item  We assume a set $\Phi$ whose elements we call \textbf{states}, one of which we single out as the \textbf{present state}.
    \item A \textbf{declarative program} is a function $f : \Phi \rightarrow \{true, false\}$, and we write $P$ for the set of all declarative programs. By an \textbf{objective truth} about a state $\phi$, we mean a declarative program $f$ such that $f(\phi) = true$.
    }
\end{itemize}
\end{definition}

\begin{definition}[implementable language] \label{d2}
\begin{itemize}{
    \item $\mathfrak{V} = \{V \subset P : V \ is \ finite\}$ is a set whose elements we call \textbf{vocabularies}, one of which\footnote{The vocabulary $\mathfrak{v}$ we single out represents the sensorimotor circuitry with which an organism enacts cognition - their brain, body, local environment and so forth.} we single out as \textbf{the vocabulary} $\mathfrak{v}$.
    \item ${L_\mathfrak{v}} = \{ l \subseteq \mathfrak{v} : \exists \phi \in \Phi \ (\forall p \in l : p(\phi) = true) \}$ is a set whose elements we call \textbf{statements}. $L_\mathfrak{v}$ follows $\Phi$ and $\mathfrak{v}$, and is called \textbf{implementable language}.
    \item  $l \in {L_\mathfrak{v}}$ is \textbf{true} iff the present state is $\phi$ and $\forall p \in l : p(\phi) = true$.
    \item The \textbf{extension of a statement} $a \in {L_\mathfrak{v}}$ is $Z_a = \{b \in {L_\mathfrak{v}} : a \subseteq b\}$.
    \item The \textbf{extension of a set of statements} $A \subseteq {L_\mathfrak{v}}$ is $Z_A = \bigcup\limits_{a \in A} Z_a$.
    }
\end{itemize}

\noindent {\normalfont(Notation)} $Z$ with a subscript is the extension of the subscript\footnote{e.g. $Z_s$ is the extension of $s$.}. 

\end{definition}

\noindent A goal can now be expressed as a statement in an implementable language. An implementable language represents sensorimotor circuitry\footnote{Mind, body, local environment etc.} with which cognition is enacted. It is not natural language, but a dyadic system with exact meaning. 
 Peircean semiosis \cite{peirce} is integrated to explain natural language. Peirce defined a symbol as a sign (E.G. the word ``pain''), a referent (E.G. the experience of pain), and an interpretant which links the two, ``determining the effect upon'' the organism. A goal arguably functions as an interpretant because it determines the effect of a situation upon an organism that pursues it \cite{bennett2022a}. 
Rather than formulate a task and then rehash the argument that a task is a symbol, we'll just formalise a symbol using the existing definition of a task \cite[definition 3]{bennett2023appendices}:

\begin{definition}[{$\mathfrak{v}$}-task]\label{d3} For a chosen $\mathfrak{v}$, a task $\alpha$ is a triple $\langle {S}_\alpha, {D}_\alpha, {M}_\alpha \rangle$, and $\Gamma_\mathfrak{v}$ is the set of all tasks given $\mathfrak{v}$. Give a task $\alpha$:\begin{itemize}{ 
    \item ${S}_\alpha \subset L_\mathfrak{v}$ is a set whose elements we call \textbf{situations} of $\alpha$.
    \item ${S_\alpha}$ has the extension $Z_{S_\alpha}$, whose elements we call \textbf{decisions} of $\alpha$. 
    \item ${D_\alpha} = \{z \in Z_{S_\alpha} : z \ is \ correct \}$ is the set of all decisions which complete $\alpha$. 
    \item ${M_\alpha} = \{l \in L_\mathfrak{v} : {Z}_{S_\alpha} \cap Z_{l} = {D_\alpha}\}$ whose elements we call \textbf{models} of $\alpha$.}
\end{itemize}
\noindent{\normalfont(Notation)} If $\omega \in \Gamma_\mathfrak{v}$, then we will use subscript $\omega$ to signify parts of $\omega$, meaning one should assume $\omega = \langle {S}_\omega, {D}_\omega, {M}_\omega \rangle$ even if that isn't written.\\

\noindent {\normalfont(How a task is completed)} Assume we've a $\mathfrak{v}$-task $\omega$ and a hypothesis $\textbf{h} \in L_\mathfrak{v}$ s.t.\begin{enumerate}{
    \item we are presented with a situation ${s} \in {S}_\omega$, and
    \item we must select a decision $z \in Z_{s} \cap Z_\textbf{h}$.
    \item If $z \in {D}_\omega$, then $z$ is correct and the task is complete. This occurs if $\textbf{h} \in {M}_\omega$.}
\end{enumerate}
\end{definition}

\begin{definition}[symbol] A task $\alpha$ is also a Peircean symbol:\begin{itemize}{ 
    \item $s \in S_\alpha$ is a \textbf{sign} of $\alpha$.
    \item $d \in D_\alpha$ is the effect of $\alpha$ upon one who perceives it. $d$ may be sensorimotor activity associated with perception, and thus a \textbf{referent}.  
    \item $m \in M_\alpha$ is an \textbf{interpretant} linking \textbf{signs} to \textbf{referents}.
    }
\end{itemize}
\end{definition}
\noindent Tasks may be divided into narrower child tasks, or merged into parent tasks.
\begin{definition}[child, parent and weakness] A symbol $\alpha$ is a child of $\omega$ if ${S}_\alpha \subset {S}_\omega$ and ${D}_\alpha \subseteq {D}_\omega$. 
This is written $\alpha \sqsubset \omega$. We call $\lvert D_\alpha \rvert$ the weakness of a symbol $\alpha$, and a parent is weaker than its children.
\end{definition}
\subsection{Extending the formalism}
The child and parent relation means a symbol is also a symbol system in that it can be subdivided into child symbols \cite{bennett2022a}. With this in mind, we can define an organism that derives symbols from its experiences, preferences and feelings.
\begin{definition}[organism] An organism $\mathfrak{o}$ is a quintuple $\langle \mathfrak{v}_\mathfrak{o}, \mathfrak{e}_\mathfrak{o}, \mathfrak{s}_\mathfrak{o}, n_\mathfrak{o}, f_\mathfrak{o} \rangle$, and the set of all such quintuples is $\mathfrak{O}$ where:
\begin{itemize}{
    \item $\mathfrak{v}_\mathfrak{o}$ is a \textbf{vocabulary} we single out as belonging to this organism\footnote{The corresponding $L_{\mathfrak{v}_\mathfrak{o}}$ is all sensorimotor activity in which the organism may engage.}.
    \item We assume a $\mathfrak{v}_\mathfrak{o}$-task $\beta$ wherein $S_\beta$ is every situation in which $\mathfrak{o}$ has made a decision, and ${D}_\beta$ contains every such decision. Given the set $\Gamma_{\mathfrak{v}_\mathfrak{o}}$ of all tasks, $\mathfrak{e}_\mathfrak{o} = \{\omega \in \Gamma_{\mathfrak{v}_\mathfrak{o}} : \omega \sqsubset \beta \}$ is a set whose members we call \textbf{experiences}.
    \item A \textbf{symbol system} $\mathfrak{s}_\mathfrak{o} = \{\alpha \in \Gamma_{\mathfrak{v}_\mathfrak{o}} : there \ exists \ \omega \in \mathfrak{e}_\mathfrak{o} \ where \ {M}_\alpha \cap {M}_\omega \neq \emptyset \}$ is a set whose members we call \textbf{symbols}. $\mathfrak{s}_\mathfrak{o}$ is the set of every task to which it is possible to generalise (see \cite[definition 5]{bennett2023appendices}) from an element of $\mathfrak{e}_\mathfrak{o}$.
    \item $n_\mathfrak{o} : \mathfrak{s}_\mathfrak{o} \rightarrow \mathbb{N}$ is a function we call \textbf{preferences}.
    \item $f_\mathfrak{o} : \mathfrak{s}_\mathfrak{o} \rightarrow \mathfrak{f}_\mathfrak{o}$ is a function, and $\mathfrak{f}_\mathfrak{o} \subset L_{\mathfrak{v}_\mathfrak{o}}$ a set whose elements we call \textbf{feelings}, being the reward, qualia etc, from which preferences arise\footnote{Note that this assumes qualia, preferences and so forth are part of physical reality, which means they are sets of declarative programs.}.
    }
\end{itemize}
\end{definition}
\noindent Each symbol in $\mathfrak{s}_\mathfrak{o}$ shares an interpretant at least one experience\footnote{A symbol system is every task to which one may generalise from one's experiences. Only finitely many symbols may be entertained. In claiming our formalism pertains to meaning in natural language we are rejecting arguments, such as those of Block and Fodor \cite{block1972}, that a human can entertain an infinity of propositions (because time and memory are assumed to be finite, which is why $\mathfrak{v}_\mathfrak{o}$ is finite).}. This is so feelings $f_\mathfrak{o}$ ascribed to symbols can be grounded in experience.
Humans are given impetus by a complex balance of feelings (reward signals, qualia etc). It is arguable that feelings eventually determine all value judgements \cite{bennettmaruyama2022a}. As Hume pointed out, one cannot derive a statement of what ought to be from a statement of what is. Feelings are an ought from which one may derive all other oughts. If meaning is about intent, then the impetus that gives rise to that intent is an intrinsic part of all meaning \cite{bennett2022b}. Intent is a goal. A goal is statement of what ought to be that one tries to make into a description of what is, by altering the world to fit with ought to be. We assume feelings are consequence of natural selection, and so explain meaning in virtue of a mechanistic process.
Each $l \in L$ represents sensorimotor activity, which from a materialist perspective includes feelings. Thus, $f_\mathfrak{o}$ is a function from symbols to sensorimotor activity. Statements and symbols ``mean something'' to the organism if the organism can ascribe feelings to them. As every symbol in $\mathfrak{s}_\mathfrak{o}$ contains an interpretant which is part of the organism's experience, the organism can ascribe feelings to all symbols on the basis of that experience.
If one is not concerned with qualia \cite{chalmers1995,boltuc2012}, then feelings may be simulated with ``reward'' functions. However, to simulate feelings that result in human-like behaviour is a more difficult proposition. 
Rather than trying to describe human-like feelings, we simplify our analysis by assuming the preferences \cite{alexander2020}  $n_\mathfrak{o}$ which are determined by experience of feelings.

\subsection{Interpretation}
The \textbf{situation at hand} $s \in L_{\mathfrak{v}_\mathfrak{o}}$ is a statement $\mathfrak{o}$ experiences as a sign and then \textbf{interprets} using $\alpha \in \mathfrak{s}_\mathfrak{o}$ s.t. $s \in S_\alpha$, to decide $d \in Z_s \cap Z_{M_\alpha}$. 
\begin{definition}[interpretation] Interpretation is a sequence of steps:
\begin{enumerate}{
    \item The situation at hand $s \in L_{\mathfrak{v}_\mathfrak{o}}$ \textbf{signifies} a symbol $\alpha \in \mathfrak{s}_\mathfrak{o}$ if $s \in S_\alpha$.
    \item $\mathfrak{s}_\mathfrak{o}^s = \{ \alpha \in \mathfrak{s}_\mathfrak{o} : s \in S_\alpha \}$ is the set of all symbols which $s$ signifies.
    \item If $\mathfrak{s}_\mathfrak{o}^s \neq \emptyset$ then $s$ \textbf{means something} to the organism in the sense that there are feelings which can be ascribed to symbols in $\mathfrak{s}_\mathfrak{o}^s$. 
    \item If $s$ means something, then $\mathfrak{o}$ uses $\alpha \in \underset{\omega \in \mathfrak{s}_\mathfrak{o}^s}{\arg\max} \ n_{\mathfrak{o}}(\omega)$ to interpret $s$.
    \item The interpretation is a decision $d \in Z_s \cap Z_{M_\alpha}$\footnote{How an organism responds to a sign that means nothing is beyond this paper's scope.}. 
    }
\end{enumerate}
\end{definition}

\section{Communication of meaning}

We develop our explanation in four parts. First, we define exactly what it means for an organism to affect and be affected by others. Second, we examine how one organism may anticipate the behaviour (by inferring the end it serves) of another or order to change how they are affected. Third, we examine how said organism may, having anticipated the behaviour of the other, intervene to manipulate the other's behaviour to their benefit (so that the now latter affects the former in a more positive way). And finally, we examine what happens when each organism is attempting to manipulate the another. Each anticipates the other's manipulation, because each anticipates the other's behaviour by inferring its intent. An organism can then attempt to deceive the other organism (continue the manipulative approach), or attempt to co-operate (communicate in good faith), a choice resembling an iterated prisoner's dilemma. 
We assume organisms make decisions based upon preferences, but preferences are not arbitrary. Feelings and thus preferences exist in virtue of natural selection, which to some extent must favour rational behaviour (to the extent that selection is significantly impacted). In this might be understood as alignment, to use AI safety terms. One's feelings are the result of alignment by genetic algorithm, and one's preferences are the result of reinforcement learning using those feelings (to determine reward). Thus we assume preferences are a balance of what is rational, and what is tolerably irrational, given the pressures of natural selection. We call this balance \textbf{reasonably performant}. 
The specifics of inductive inference are beyond the scope of this paper, however definitions and formal proofs pertaining to inductive inference from child to parent tasks are included in the appendix \cite{bennett2023appendices}. The necessary inductive capabilities are assumed with being reasonably performant.

\subsection{Ascribing intent}
\begin{definition}[affect] To affect an organism $\mathfrak{o}$ is to cause it to make a different decision than it otherwise would have. $\mathfrak{k}$ affects $\mathfrak{o}$ if $\mathfrak{o}$ would have made a decision $d$, but as a result of a decision $c$ made by $\mathfrak{k}$, $\mathfrak{o}$ makes decision $g \neq d$.

\end{definition}
Let $\mathfrak{k}$ and $\mathfrak{o}$ be organisms. If $\mathfrak{k}$ affects $\mathfrak{o}$, and assuming $\mathfrak{v}_\mathfrak{o}$ is sufficient to allow $\mathfrak{o}$ to distinguish when it is affected by $\mathfrak{k}$ from when it is not (meaning all else being equal $\mathfrak{k}$'s interventions are distinguishable by the presence of an identity (see appendices), then there exists experience $\zeta_\mathfrak{o}^\mathfrak{k} \in \mathfrak{e}_\mathfrak{o}$ such that $d \in D_{\zeta_\mathfrak{o}^\mathfrak{k}}$ if $\mathfrak{o}$ is affected by $\mathfrak{k}$. $\zeta_\mathfrak{o}^\mathfrak{k}$ is an ostensive definition \cite{gupta2021} of $\mathfrak{k}$'s intent (meaning it is a child task from which we may infer the parent representing $\mathfrak{k}$'s most likely intent and thus future behaviour) \cite{bennett2022a}.
In the absence of more information, the symbol most likely to represent $\mathfrak{k}'s$ intent is the weakest \cite{bennett2022a}, meaning  $\alpha \in \mathfrak{s}_\mathfrak{o}$ s.t. $\lvert Z_\alpha \rvert$ is maximised. However, because $\mathfrak{o}$ assumes $\mathfrak{k}$ has similar feelings and preferences \cite{bennett2022a,bennettmaruyama2022a}\footnote{Members of a species tend to have similar feelings, experiences and thus preferences.} $n_\mathfrak{o}$ is an approximation of what $\mathfrak{k}$ will do. Accordingly the symbol most likely to represent $\mathfrak{k}$'s intent would be the ``weakest'' of goals preferred by $\mathfrak{o}$ which, if pursued by $\mathfrak{k}$, would explain why $\mathfrak{k}$ has affected $\mathfrak{o}$ as it has. This is $\gamma_\mathfrak{o}^\mathfrak{k} $ s.t. 
        $$\gamma_\mathfrak{o}^\mathfrak{k} \in \underset{\alpha \in \mathfrak{K}}{\arg\max} \ \lvert Z_\alpha \rvert \text{ s.t. } \mathfrak{K} = \underset{\alpha \in \Gamma_\mathfrak{o}^\mathfrak{k}}{\arg\max} \ n_\mathfrak{o}(\alpha) \text{ and } \Gamma_\mathfrak{o}^\mathfrak{k} = \{\omega \in \Gamma_{\mathfrak{v}_\mathfrak{o}} : M_{\zeta_\mathfrak{o}^\mathfrak{k}} \cap M_\omega \neq \emptyset \} $$ 

\subsection{From manipulation to meaningful communication}
We've explained inference of intent in counterfactual terms, answering ``if places were exchanged, what would cause $\mathfrak{o}$ to act like $\mathfrak{k}$?''. Intent here is ``what is $\mathfrak{k}$ trying to achieve by affecting $\mathfrak{o}$'', rather than just ``what is $\mathfrak{k}$ trying to achieve''.

\subsubsection{Manipulation:} Because it is reasonably performant, $\mathfrak{o}$ infers the intent of an organism $\mathfrak{k}$ that affects $\mathfrak{o}$, in order to plan ahead and ensure its own needs will be met. However $\mathfrak{o}$ can go a step further. It can also attempt to influence what $\mathfrak{k}$ will do. If being reasonably performant requires $\mathfrak{o}$ infer $\mathfrak{k}$'s intent because $\mathfrak{k}$ affects $\mathfrak{o}$, then it may also require $\mathfrak{o}$ affect $\mathfrak{k}$ to the extent that doing so will benefit $\mathfrak{o}$.

\subsubsection{Communication:} If both $\mathfrak{o}$ and $\mathfrak{k}$ are reasonably performant, each may attempt to manipulate the other. Ascribing intent to one another's behaviour in order to manipulate, each must anticipate the other's manipulative intent. Subsequently each organism must go yet another step further and account for how its own manipulative intent will be perceived by the other. As in a sort of iterated prisoner's dilemma, the rational choice may then be to co-operate. Because each symbol represents a goal it defines a limited context for co-operation; so two organisms might simultaneously co-operate in pursuit of one goal while competing in pursuit of another (E.G. two dogs may co-operate to hunt while competing for a mate). 
If there is sufficient profit in affecting another's behaviour, then knowing one's own intent is perceived by that other and that the other will change its behaviour in response to one's changed intent, it makes sense to actually change one's own intent in order to affect the other. This bears out experimentally in reinforcement learning with extended environments \cite{alexander2022}.
The rational choice may then be to \textit{have} co-operative intent, assuming $\mathfrak{k}$ can perceive $\mathfrak{o}$'s intent correctly, and that  $\mathfrak{k}$ will reciprocate in kind.
For a population of reasonably performant organisms, induction (see \cite{bennett2023appendices}) with co-operative intent would favour symbols that mean (functionally) similar things to all members of the population. Repeated interactions would give rise to signalling conventions we might call language.

\subsubsection{Meaning:} Let us re-frame these ideas using the example from the introduction. We'll say two symbols $\alpha \in \mathfrak{s}_\mathfrak{k}$ and $\omega \in \mathfrak{s}_\mathfrak{o}$ are roughly equivalent (written $\alpha \approx \omega$) to mean feelings, experiences and thus preferences associated with a symbol are in some sense the same for two organisms (meaning if $\alpha \approx \omega$ then $f_\mathfrak{k}(\alpha) \approx f_\mathfrak{o}(\omega)$ etc). In other words we're suggesting it must be possible to measure the similarity between symbols in terms of feelings, experiences and thus preferences, and so we can assert a threshold beyond which two symbols are roughly equivalent.
\begin{quote}{
$\mathfrak{k}$ means $\alpha \in \mathfrak{s}_\mathfrak{k}$ by deciding $u$ and affecting $\mathfrak{o}$ iff $\mathfrak{k}$ intends in deciding $u$: \\
1. that $\mathfrak{o}$ interpets the situation at hand with $\omega \in \mathfrak{s}_\mathfrak{o}$ s.t. $\omega \approx \alpha$, \\
2. $\mathfrak{o}$ recognize this intention, for example by predicting it according to $$\gamma_\mathfrak{o}^\mathfrak{k} \in \underset{\alpha \in \mathfrak{K}}{\arg\max} \ \lvert Z_\alpha \rvert \text{ s.t. } \mathfrak{K} = \underset{\alpha \in \Gamma_\mathfrak{o}^\mathfrak{k}}{\arg\max} \ n_\mathfrak{o}(\alpha), \Gamma_\mathfrak{o}^\mathfrak{k} = \{\omega \in \Gamma_{\mathfrak{v}_\mathfrak{o}} : M_{\zeta_\mathfrak{o}^\mathfrak{k}} \cap M_\omega \neq \emptyset \}$$
3. and (1) occur on the basis of (2), because $\mathfrak{k}$ intends to co-operate and so will interpret the situation at hand using what it believes of $\mathfrak{o}$'s intent. }
\end{quote}

\noindent The above pertains to co-operation. To comprehend meaning:

\begin{enumerate}{
    \item Organisms must be able to \textbf{affect one another}.
    \item Organisms must have similar \textbf{feelings}, and
    \item similar \textbf{experiences}, so $\mathfrak{s}_\mathfrak{o}$ and $\mathfrak{s}_\mathfrak{k}$ contain roughly equivalent symbols.
    \item Similar \textbf{preferences} then inform the correct inference of intent.
    \item Finally, all this assumes organisms are \textbf{reasonably performant}.
    }
\end{enumerate}

\section{Talking to a machine}
An LLM is trained to mimic human preferences. However, an LLM is not given impetus by feelings, and so cannot entertain roughly equivalent symbols. This is not to say we cannot reverse engineer the complex balance of human-like feelings, merely that we have not. If an LLM has impetus, it is to be found in our prompts. It is reminiscent of a mirror test, which is a means of determining whether animals are self aware. For example, a cat seeing itself in the mirror may attack its reflection, not realising what it sees is itself. In an LLM we face a mirror test of our own, but instead of light it reflects our own written language back at us. We ascribe motives and feelings to that language because we have evolved to infer the intent of organisms compelled by feelings \cite{bennett2022a}. An LLM hijacks what we use to understand one another (that we assume others are motivated by similar feelings \cite{bennettmaruyama2022a}). We've a history of ascribing feelings and agency to things possessed of neither. In the 1970s, a chatbot named ELIZA made headlines as its users attributed feelings and motives to its words \cite{weizenbaum1976}. Like ELIZA, today's LLMs not only do not mean what we think they mean by what they say, but do not mean anything at all. This is not an indictment of LLMs trained to mimic human preferences. The meaning we ascribe to their behaviour can be useful, even if that behaviour was not intended to mean anything. 
\subsubsection{The Hall of Mirrors:} 
Even if we approximate human feelings, an LLM like ChatGPT is not reasonably performant. It is maladaptive, requiring an abundance of training data. This may be because training does not optimise for a weak representation, but settles for any function fitting the data\footnote{Albeit with some preference for simplicity imparted by regularisation.} \cite{bennett2022a}. Returning to mirror analogies, imagine a hall of mirrors reflecting an object from different angles. A weak or simple representation would be one symbol $\alpha \in \mathfrak{s}_\mathfrak{o}$ representing the object, which is then interpreted from the perspectives $a, b, c, d \in S_\alpha$ of each mirror. A needlessly convoluted representation of the same would instead interpret $a, b, c$ and $d$ using different symbols. These would be $\alpha$’s children $\omega, \gamma, \delta, \sigma\sqsubset \alpha$ such that $a \in S_\omega,b \in S_\gamma,c \in S_\delta,d \in S_\sigma$. This latter representation fails to exploit what is common between perspectives, which might allow it to generalise \cite{bennett2022a} to new perspectives.
That an LLM may not learn sufficiently weak representations seems consistent with their flaws. One well documented example of this is how an LLM may convincingly mimic yet fail to understand arithmetic \cite{floridi2020}, but such flaws may more subtly manifest elsewhere. For example, when we queried Bing Chat (on the $2^{nd}$ of February 2023 \cite[p.11]{bennett2023appendices}) with the name and location of a relatively unknown individual who had several professions and hobbies mentioned on different sites, Bing concluded that different people with this name lived in the area, each one having a different hobby or profession.

\subsubsection{Incomprehensibility:} 
If we are to build machines that mean what we think they mean by what they say, then we must emulate human feelings and experiences. 
It is interesting to consider where this may lead. If we do not get the balance of feelings quite right, we might create an organism that means what it says, but whose meanings may be partially or utterly incomprehensible to us because the resulting preferences are unaligned with ours. In the introduction we mentioned ideas on which this paper was founded were used to relate the Fermi paradox to control of and communication with an AGI \cite{bennett2022b}. We can extend that notion. 
Assume we are affected by an organism. If the events befalling us are set in motion by preferences entirely unlike our own, then we would fail to ascribe the correct intent to the organism. We may fail entirely to realise there is an organism, or may ascribe many different intents as in the hall of mirrors analogy. Furthermore, $\mathfrak{v}_\mathfrak{o}$ determines what can or cannot be comprehended by an organism (see appendices). $\mathfrak{v}_\mathfrak{o}$ may contain nothing akin to the contents of $\mathfrak{v}_\mathfrak{k}$, making $\mathfrak{o}$ incapable of representing and thus comprehending $\mathfrak{k}$'s intent. 

\subsubsection{Conclusion:} We have extended a formalism of artificial general intelligence, connecting bottom up computation to top down notions of meaning. This is significant not just to AGI but to wider debates surrounding language, meaning and the linguistic turn. While we focused on Gricean notions of meaning due to publication constraints, the formalism is by no means limited to that. For example, the logical truth conditional meaning of statements is in their extension. We have described the process by which meaningful communication can take place and the prerequisites thereof. We conclude that human-like feelings and weak representations should give us systems that comprehend and intend meaning. 

\printbibliography

\end{document}